\documentclass{article}

\usepackage{subcaption}

\usepackage[final]{corl_2020} 

\usepackage{booktabs}
\usepackage{graphicx}
\usepackage[toc,page]{appendix}

\usepackage{amsmath}
\usepackage{amssymb}
\usepackage{algorithm,algorithmic}
\usepackage{url}
\usepackage{color}
\usepackage{caption}

\def\shownotes{0}  
\ifnum\shownotes=1
\newcommand{\authnote}[2]{{$\ll$\textsf{\footnotesize #1 notes: #2}$\gg$}}
\else
\newcommand{\authnote}[2]{}
\fi

\author{
    Mohit Sharma\\
    Robotics Institute\\
    Carnegie Mellon University \\
    \texttt{mohits1@andrew.cmu.edu} \\
    \And
    Oliver Kroemer \\ 
    Robotics Institute\\
    Carnegie Mellon University \\
    \texttt{okroemer@andrew.cmu.edu} \\
}

\newcommand{\todo}[1]{}
\renewcommand{\todo}[1]{{\color{red}Todo: {#1}}}

\definecolor{alizarin}{rgb}{0.82, 0.1, 0.26}
\definecolor{table_color_train}{rgb}{0.682, 0.612, 0.271}
\definecolor{table_color_test}{rgb}{0.376, 0.451, 0.694}
\definecolor{mypink3}{cmyk}{0, 0.7808, 0.4429, 0.1412}
\definecolor{mygray}{gray}{0.6}

\begin{document}

\title{Relational Learning for Skill Preconditions}



%

\maketitle

\begin{abstract}
To determine if a skill can be executed in any given environment, a robot needs to learn the preconditions for the skill.
As robots begin to operate in dynamic and unstructured environments, precondition models will need to generalize to variable number of objects with different shapes and sizes.
In this work, we focus on learning precondition models for manipulation skills in unconstrained environments.
Our work is motivated by the intuition that many complex manipulation tasks, with multiple objects, can be simplified by focusing on less complex pairwise object relations.
We propose an object-relation model that learns continuous representations for these pairwise object relations. 
Our object-relation model is trained completely in simulation, and once learned, is used by a separate precondition model to predict skill preconditions for real world tasks.
We evaluate our precondition model on $3$ different manipulation tasks: sweeping, cutting, and unstacking. 
We show that our approach leads to significant improvements in predicting preconditions for all $3$ tasks, across objects of different shapes and sizes.

\end{abstract}


\section{Introduction}




Skill \emph{preconditions} are necessary for a robot to know when a given skill can be executed across different situations. For robot manipulators to operate in complex and unstructured environments, we require precondition models to be able to adapt to variable number and type of objects in the scene, \emph{e.g.}, preconditions for food cutting should generalize between apples and carrots.
Additionally, since collecting large amounts of real world data for every skill is impractical, we also require precondition models to learn from few samples.
To achieve the above requirements, we observe that many complex manipulation tasks often require specific relations between a number of objects to be valid. Thus, these tasks can often be simplified by decomposing them into less complex object interactions. 
We therefore focus on modeling pairwise relations between objects.
Moreover, these object relations are also often shared across multiple tasks. Thus, we also aim to learn a common representation for these relations, which can be directly used by our precondition models.

\begin{figure}[t]
    \centering
    \includegraphics[width=1.0\linewidth]{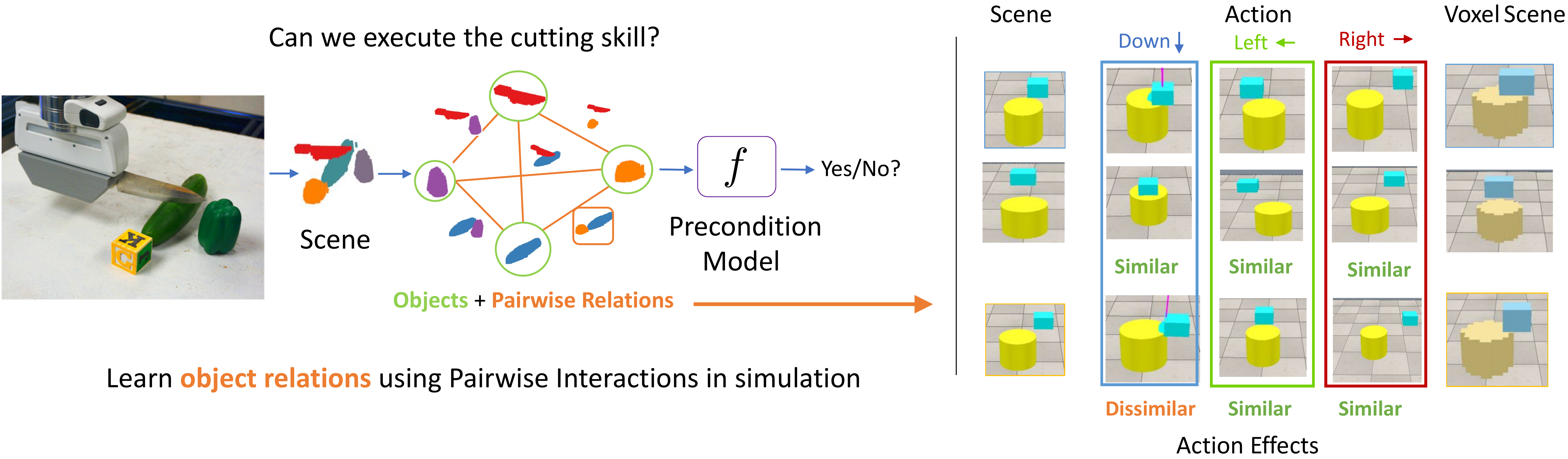}
    \caption{\footnotesize{Overview of the overall precondition learning process. We learn precondition model by decomposing a scene into its objects and relations. To learn continuous representations for object relations we utilize simulations to observe how objects interact with each other.}}
    \label{fig:teaser_image}
    \vspace{-4mm}
\end{figure}

Our work uses a compositional approach for precondition learning.
We learn precondition models from a few different example scenes. Each scene is decomposed into its constituent objects and their relations as shown in Figure~\ref{fig:teaser_image}. Learning precondition models from such a structured representation requires us to infer object identities and pairwise object relations.
While object labels can be found using state-of-the-art algorithms \cite{he2017maskrcnn}, we focus on learning representations for object pair relations.
Most of the prior work uses discrete representations for object relations \cite{rosman2011learning, fichtl2014learning, zampogiannis2015learning}. These discrete relations are mostly fixed and often insufficient to distinguish between similar scenes. 
For instance, in Figure~\ref{fig:teaser_image}, even though both relations might be classified as ``above'' the semantics of each scene are very different. 
Although recent approaches have looked at learning continuous representations for object relations \cite{mees2017metric, jund2018optimization},
these approaches use a pre-defined set of relations. 
They further assume access to binary labels which allows them to cluster scenes with similar relations together.

Instead of assuming a fixed set of relations we propose a novel data-driven approach to learn continuous representations for object relations. We leverage simulations to allow for unconstrained interactions between object pairs.
Using simulations provides us access to a large set of labeled data underlying these interactions, such as object contacts, pose changes.
We use these interaction effects to create a contrastive loss formulation that allows us to learn continuous representations for object relations.
Our main contributions include 
1) We propose a novel approach that leverages simulations to learn a continuous representation for object relations.
2) Our approach is grounded in object interactions and hence does not assume a fixed set of object relations.
3) We show that our continuous representations, which are learned via simulation, can be directly used for precondition learning of real world manipulation tasks, without requiring any specific sim2real adaptation.

\section{Related Works}
Learning skill preconditions is closely related with symbol learning. Symbolic language formulations explicitly include pre-conditions and post-conditions \cite{mcdermott1998pddl}. 
Thus previous approaches have looked at inferring discrete symbols and predicates from continuous states. 
There have been two sets of approaches in this regard.
At one end, previous works have looked into classifying continuous states into pre-defined symbolic forms \cite{paxton2016want, kaelbling2013integrated, beetz2010cram, fichtl2014learning}. Alternately, other works aim to discover these discrete forms by methods such as clustering \cite{kulick2013active, jetchev2013learning, pasula2007learning, rosman2011learning}.
Some of the above approaches have also looked into learning discrete representations for object relations. For instance, \cite{fichtl2014learning} uses histogram features extracted from point clouds to classify object relations into discrete categories. 
In \cite{rosman2011learning}, the authors construct a graph of contact points between two objects, which is used to classify object relations using supervised learning.
While \cite{zampogiannis2015learning} learns discrete spatial relations by geometrically dividing the space around an object into a fixed set of regions. 
In contrast to discrete symbolic forms, we focus on learning continuous representations for object relations.
Also, instead of solely focusing on either geometric relations \cite{zampogiannis2015learning} or contact points \cite{rosman2011learning}, our approach is grounded in a robot's interaction with pairs of objects.
This allows us to learn continuous relational representations by leveraging spatial geometry, contact information, and object-interaction dynamics all together.

Recent work has also focused on learning continuous representations for object relations \cite{mees2017metric, jund2018optimization}.
Both \cite{mees2017metric, jund2018optimization} learn these representations from a dataset of $ N \sim 500$ scenes.
Each scene has a pair of objects, which are sampled from a set of known object templates.
The object pair arrangement is sampled from a pre-defined set of relations and the aim is to generalize these arrangements to new sets of objects. The dataset is manually labelled with binary similarity value for each scene pair \cite{mees2017metric}.
Based on this dataset \cite{mees2017metric} proposes a distance metric to extract similar scenes.
While \cite{jund2018optimization} uses contrastive learning to learn this distance metric.
Although \cite{mees2017metric, jund2018optimization} learn continuous relationas, the dataset used to learn these consists of a pre-defined set of discrete relations only. 
Additionally, using only binary labels to distinguish between scenes is limiting, \emph{e.g.} scenes where object A is on the left of object B are labeled similarly irrespective of the distance between two objects.
By contrast, rather than manually labelling scenes, we leverage simulations to generate large amount of object pair interactions. We group similar scenes together based on effets of these interaction. Since these effects have continuous values, our model is able to learn richer relational representations.
Further, since these interactions are randomly generated, we do not assume any predefined set of relations. 

Previous work \cite{kroemer2016learning} has also looked into precondition learning using objects, parts and their interactions. In \cite{kroemer2016learning} these are referred to as scene elements and represented by their mean 3D position. All scene elements are then used together for spatial precondition learning using random forests.
However, using position features alone is severely limiting since many manipulations tasks depend on object sizes, orientation, geometry, and specific contact distributions.

\section{Approach}

We learn preconditions for manipulation skills by decomposing a scene $S$ into its constituent objects $(o_1, o_2, o_3, \ldots)$ and their continuous relations $(r(o_1, o_2), r(o_1, o_3),  \ldots)$. 
We assume the scene decomposition is already known and object identities can be determined if required.
To infer object relations $r(o_i, o_j)$, we learn a function $f_{\text{rel}}: \mathbb{R}^{C\times L \times W \times H}  \to \mathbb{R}^K$, which uses a 3D voxel based perceptual input, 
where $C$ is the number of channels and $L, W, H$ are the length, width and height of the scene, and outputs a continuous relational embedding of size $K$. 
This perceptual input consists of a pair of objects in the scene.
To learn $f_\text{rel}$, we leverage simulations to generate a large set of pairwise object interactions and use their observed effects as our learning signal.
More specifically, we assume the existence of a set of simple perturbation actions that allow a robot to move an object around. These actions let the robot freely interact with pairs of objects.
The use of simulation also gives us access to the underlying effects of these interactions \emph{e.g.}, change in pose, object contacts and normals, and force-torque values.
We utilize these observed effects to create a contrastive learning formulation which groups scenes together based on the similarity of these interaction effects.


Once we learn $f_\text{rel}$, we use it to extract object relations $r(o_i, o_j) = f_\text{rel}(o_i, o_j)$ for all object pairs in the scene. These extracted relations are then used as input to our precondition learning model, which is trained directly on real world manipulation task data. 
While training the precondition model, we do not fine-tune the object relation model $f_\text{rel}$.
We show that $f_\text{rel}$ trained in simulation can be directly transferred to the real world without requiring any sim2real adaptation. 


\subsection{Generating Pairwise Interactions In Simulation}

Pairwise object relations are closely intertwined with their potential interactions. For instance, in Figure~\ref{fig:teaser_image} (right) if an object is gently placed on another object it will either stay in place or it might tilt over and fall. Although both of these scenes might classify a discrete object relation as being ``above", the exact semantics of these relations are quite different.
To generate object pair interactions we use the V-REP simulator \cite{rohmer2013v} with the underlying Bullet physics engine (v2.83) \cite{coumans2013bullet}.
For each interaction, we initially create a scene with two objects $o_i$ and $o_j$. 
Each object's shape is chosen from a fixed set of primitive shapes including cuboids, cylinders and spheres.
We generate a voxel representation for this scene using an object-centric approach, \emph{i.e.}, the reference frame of the scene is centered on one of the objects, referred to as the anchor object ($o_i$), while $o_j$ is the the referrant object. Figure~\ref{fig:teaser_image} (right column) shows an example of such a voxel based representation.
To observe the interaction effects of these two objects we keep the anchor object static.

Since objects in the real world occur in varying spatial locations we do not assume any fixed object positions. However, since the scene reference frame is centered on the anchor object we do not need to set its position and orientation.
To position the referrant object we sample a location $(x, y, z)$ around the anchor object such that the referrant object is less than $0.5$ meter away from the anchor. 
To set the orientation we rotate the object around the z-axis between $(-\pi/6, \pi/6)$ radians.

To create object pair interactions, we apply local perturbations to the referrant object. These local perturbations move the referrant object in the Euclidean space. Figure~\ref{fig:teaser_image} (right column) shows three different instances of local perturbations.
Formally, these perturbation can be expressed in polar format as $(r, \theta, \phi)$, where $r$ is the action magnitude and $\theta, \phi$, represents the action direction. 
We sample action directions that are axes aligned with the objects reference frame as well as along the diagonals of each axis pair. 
We use two different strategies to sample action magnitude.
First, we use a fixed action magnitude sampled from $\left[5cm, 20cm\right]$. Additionally, we also use adaptive actions, wherein the action magnitude is set as the distance between the anchor and the referrant object centers. These adaptive actions ensure that objects will interact even if the distance between them is greater than the maximum fixed action magnitude.
To perform local perturbation actions, we add a virtual robot with a spring-damper system.

Given the data generation process, we generate $N\sim100,000$ pairwise object scenes. 
For each scene, we save the voxel representation for the anchor and referrant objects. 
Additionally, for each action we also record the distance moved (in meters) by the referrant object $(\Delta p)$, as well as the change in orientation (in radians) of the referrant object $(\Delta \theta)$ when it interacts with the anchor object. 
For scenes which involve contacts between objects we also record the contact position and the contact normals. These contact positions are recorded for the anchor object relative to its frame of reference.
More specific details can be found in the supplementary material.

\subsection{Learning Object Relations}

We use pairwise object interactions to learn our object relation model $f_\text{rel}$.
The input to $f_\text{rel}$ is 3D voxel representation of an object pair and it outputs a continuous low dimensional relation embedding. 
This formulation does not require an action as an input and hence $f_\text{rel}$ can be used directly at test time. 
Our main insight to learn $f_{\text{rel}}$ is based on the fact that the same actions on similar scenes should result in similar effects. These action effects refer to change in properties such as object position, orientation, contacts.
We use these action effects to train $f_\text{rel}$ using two different sets of losses.
First, we take a metric learning approach and use the action effects to create contrastive losses that group similar scenes together and away from dissimilar ones.
Second, we combine the output relation embedding with the action representation to directly predict the action effects. 

\begin{figure}[t]
    \centering
    \begin{subfigure}{0.52\textwidth}
        \centering
        \includegraphics[width=1.0\textwidth]{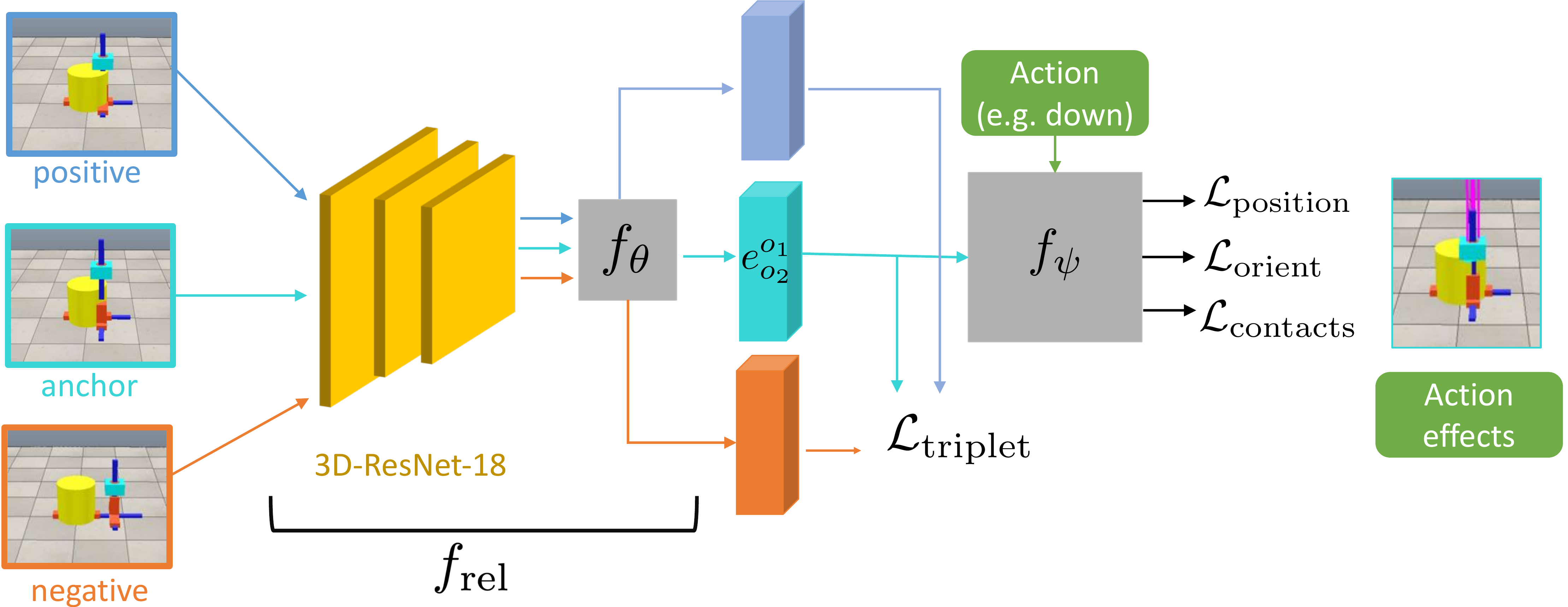}
    \end{subfigure}
    \hspace{0.2cm}
    \begin{subfigure}{0.42\textwidth}
        \centering
        \includegraphics[width=\linewidth]{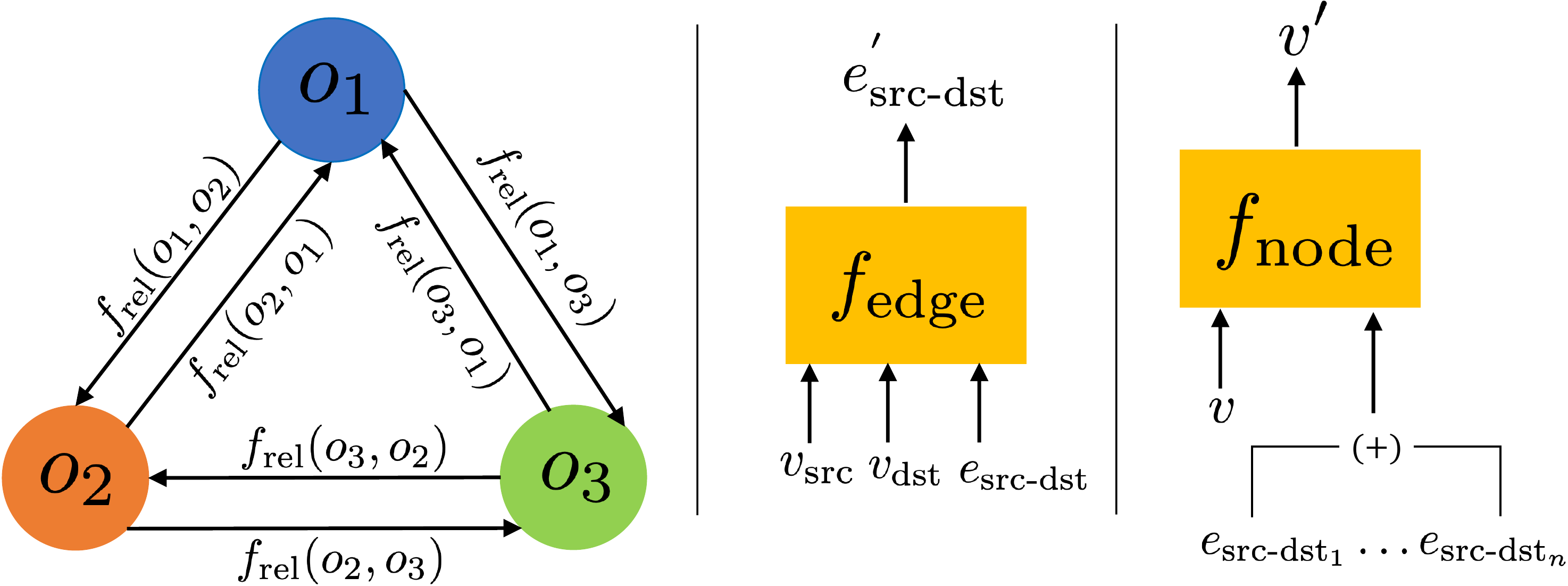}
    \end{subfigure}
    \caption{\footnotesize{\emph{Left:} Overall architecture to learn object relation model $f_\text{rel}$. \emph{Right:} Precondition learning architecture based on GNNs, shows graph model at initialization and the graph edge and node models.
    }}
    \label{fig:all_archs}
    \vspace{-3mm}
\end{figure}

\textbf{Model: }
Figure~\ref{fig:all_archs} (Left) shows the architecture of our model. We use a ResNet \cite{he2016deep} based architecture as our backbone model to process the 3D voxel input.
We use \cite{hara3dcnns} to adapt the ResNet model to operate on 3D input.
The output of the model is projected down to $\mathbb{R}^{K=256}$ using a linear projection.
We refer to the output of the linear projection as the relation embedding $r(o_i, o_j) = f_\text{rel}(X)$, where $o_i$, $o_j$ denotes the anchor and referrant object respectively and $X$ is the perceptual input with the object pair.
The projected relation embedding $r(o_i, o_j)$ is then concatenated with the action input $a$. We create $a$ by combing the action vector of size $\mathbb{R}^3$ with a one-hot label of size $\mathbb{R}^2$, this label is used to differentiate between fixed and adaptive perturbation actions.
This combined representation is then used to predict a set of outputs which correspond to the observed effects of this action. 

\textbf{Loss: }
We learn $f_\text{rel}$ using contrastive losses, which have been used to learn continuous embeddings from high dimensional data \cite{schroff2015facenet, jund2018optimization, hadsell2006dimensionality, oh2016deep}.
Contrastive losses are useful for our formulation since they compare sample similarity directly in the representation space.
Amongst the multiple contrastive loss formulations we use the triplet loss \cite{schroff2015facenet}.
Formally, the triplet loss requires an anchor $(a)$ and positive $(p)$ - negative $(n)$ pairs. The aim is to keep the anchor and positive samples together while pushing the negative sample away based on a margin, $L_{\text{triplet}} = (d_{ap} - d_{an} + \gamma)_{+}$ where $d_{ap}$ and $d_{an}$ are the squared euclidean distance between the anchor and positive embedding, and anchor and negative embedding respectively. $\gamma$ is a constant margin between similar and dissimilar pairs. 

We utilize the observed effects of the perturbation actions to get positive and negative samples for the anchor scene.
Specifically, scenes where all local perturbations lead to similar effects are more similar than other scenes.
Based on this we add triplet loss for both position $(\Delta p)$ and orientation changes $(\Delta \theta)$ separately. 
To get positive samples for $(\Delta \theta)$, we compare all action effects using $| \Delta \theta^i_{\text{1}} - \Delta \theta^i_{\text{2}} | < \Delta \theta_{\text{sim}}$, where $\Delta \theta^i_{\text{1}}$, and $\Delta \theta^i_{\text{2}}$ are the effects of $i$'th local perturbation action on two different scenes.
To achieve the same for position changes, instead of using $\Delta p$ directly we use the ratio of observed change in position to the desired change in position $\Delta p^r = \frac{\Delta p^{\text{observed}}}{\Delta p^{\text{desired}}},$
where $\Delta p^{\text{observed}}$ is the observed change in referrant object center position, and $\Delta p^{\text{desired}}$ is the desired change \emph{i.e.} action vector. 
We set $\Delta p^{\text{desired}}$ to 1 for normal actions, while for adaptive actions it is set to the voxel distance between objects centers. 
This is required to allow adaptive actions to be compared against each other since these actions result in different $\Delta p^{\text{desired}}$ values across different scenes.
We sample negative scenes by using $| \Delta \theta^i_{\text{1}} - \Delta \theta^i_{\text{2}} | > \Delta \theta_{\text{diff}}$, where we set 
$\Delta \theta_{\text{diff}} = \Delta \theta_{\text{sim}} + \epsilon $, where $\epsilon$ is a small value set manually.
We include implementation details, including the different hyper-parameters, for the contrastive loss, in the supplementary material.
In addition to contrastive losses, we also use the following prediction losses to train $f_\text{rel}$: \newline
\noindent\textbf{Predicting object positions:} We use $f_\text{rel}$ to predict the change in object position $\Delta p$ for the referrant object. However, since we use voxels for our scene representation we lose the granularity of continuous movement in the euclidean space. Hence, instead of predicting the raw values of change in position we predict the voxel displacement of the referrant object's origin $\Delta p^{v}$. 
We add a mean-squared loss on this prediction, $L_{\text{pos}} = ||\Delta p^{v}_{\text{pred}} - \Delta p^{v}_{\text{gt}}||_2$ 

\noindent\textbf{Predicting object orientations:} We also predict the change in orientation $\Delta \theta$ of the referrant object. Since the object is attached to a robot and moved controllably the magnitude of $\Delta \theta$ is usually low $\Delta \theta \sim 0.1$. Hence, instead of using $L_2$ loss we use $L_1$ loss, $L_{\text{orient}} = ||\Delta \theta_{\text{pred}} - \Delta \theta_{\text{gt}}||_1$

\noindent\textbf{Predicting contact distributions:} We also predict the contact distributions between interacting objects. Given the contact data from simulation we fit a Gaussian model to predict the mean $(x, y, z)$ of the contact point distribution. We then add the following loss, $L_{\text{contact}} = ||(\mu_{\text{pred}} - \mu_{\text{gt}}||_2$.



\subsection{Learning Precondition Models}

Once we learn the object relation model $f_\text{rel}$ from pairwise object interaction data we use it for downstream real-world precondition learning tasks. 
Our precondition learning model is based on deep neural network architectures which can operate on a set of inputs which are provided by the learned $f_\text{rel}$ model.
Formally, given a scene $S$ with objects $S=\{o_1, o_2, \ldots , o_N\}$. We use $f_\text{rel}$ to get object relation embeddings for all $N\times(N-1)$ object pairs $f_\text{rel}(o_i, o_j)$, where $o_i$ and $o_j$ are the anchor and referrant object.
For our precondition learning model we use two different network architectures \emph{i.e.} Relational Networks (RNs) and Graph Neural Networks (GNNs). RNs have been used to reason about objects and their relations \cite{santoro2017simple}. GNNs are more general neural network architectures 
where the input can be represented as a graph $G = (V, E)$, with vertices $V$ and edges $E$ \cite{scarselli2008graph, kipf2016semi, battaglia2018relational}. 
Figure~\ref{fig:all_archs} (Right) shows our formulation of a scene $S$ as a GNN. Specifically, we represent each object as a separate node $v_i \equiv o_i$ and add bi-directional edges between every node pair. At initialization, we add $f_\text{rel}(o_i, o_j)$ as the edge information for edge $e_{ij} = (v_i, v_j)$ 
Unless explicitly specified, we do not add any node information to the vertices. To allow for better inference, we stack two layers of our graph network architecture. The output of the final graph layer is classified using the sum of all the node and edge embeddings. More details are presented in the supplementary material.

\begin{figure}[t]
    \centering
    \begin{subfigure}{0.44\linewidth}
        \centering
        \includegraphics[width=0.9\textwidth]{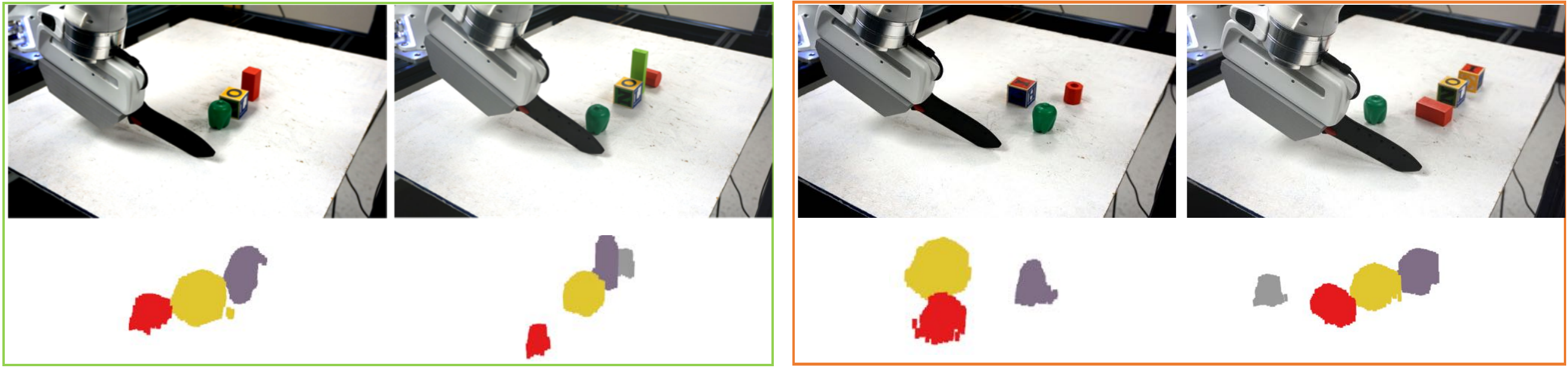}
    \end{subfigure}
    \begin{subfigure}{0.52\linewidth}
        \centering
        \includegraphics[width=0.9\linewidth]{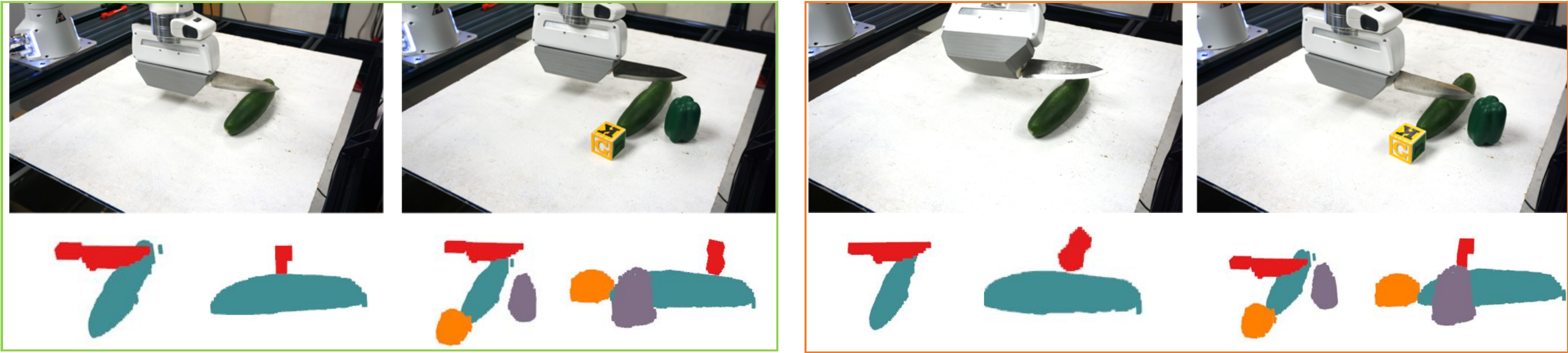}
    \end{subfigure}
    \vspace{-2mm}
    \caption{\small{\emph{Left:} Example scenes with 3 and 4 objects used for the objects in line task. \emph{Right:} Example scenes with 0 and 2 distractor objects for learning food cutting preconditions.}}
    \label{fig:data_in_line_cut_food_examples}
    \vspace{-4mm}
\end{figure}

\section{Experiments}
\vspace{-2mm}
Our aim is to investigate the following questions,
1) Given limited amount of precondition training data, how effective are the embeddings learned by our object relation model for precondition learning?
2) Does a structured representation of a scene as objects and relations
help in precondition learning for scenes with variable number of objects?
and 
3) How effectively do the precondition models generalize to objects with different shapes and sizes than the train set?
To verify these we use three different manipulation tasks: sweeping objects in a line, food cutting, and 3D block unstacking. 
Moreover, to show the effectiveness of our learned embeddings, we do not finetune the object relation model $f_\text{rel}$ during precondition learning.

\textbf{Metrics:} Since precondition learning is a binary prediction problem we compare models using F1 and weighted F1 scores. F1 score is the harmonic mean of precision and recall with higher values (max: 1.0) indicating better models and wt. F1-score is used to account for data imbalance.

\textbf{Baselines:}
We compare our approach against both learning and non-learning baseline methods.
Among learning based methods we initially use models that use 3D scene input and output the binary precondition.
We use multiple baseline architectures to represent this learned model. 
First, similar to our object relation model, we use a 3D ResNet-18 model with the output being projected down using multiple linear layers to predict the precondition output. To increase the representation capacity of this baseline model we also evaluate a ResNet-34 based model which is adapted in the same way as \cite{hara3dcnns}.
Given the limited amount of real world data, we also compare against a smaller VGG \cite{simonyan2014very} based model with 3D convolutions. 

To verify the effectiveness of continuous relations we also use a discrete relation baseline, wherein instead of continuous relations we use discrete relations. We refer to this as \emph{DiscreteRel}.
Finally, in addition to learning based methods we also evaluate another approach which utilizes the simulator to verify the preconditions. In this approach we transfer the 3D representation of a real world scene into the simulator. Once the scene has been transferred we simulate the task to verify the preconditions. We refer to this baseline as \emph{real2sim}.
See appendix for training and implementation details.
\begin{table}[]
\centering
\resizebox{0.9\textwidth}{!}{%
\begin{tabular}{@{}lllllllllll@{}}
\toprule
 & \multicolumn{4}{c}{Sweep Objects In Line} & \multicolumn{1}{l|}{} &  & \multicolumn{4}{c}{Food Cutting} \\ \midrule
Model & \begin{tabular}[c]{@{}l@{}}Train Set\\ (objects)\end{tabular} & \begin{tabular}[c]{@{}l@{}}Test Set\\ (objects)\end{tabular} & F1 & Wt-F1 & \multicolumn{1}{l|}{} &  & \begin{tabular}[c]{@{}l@{}}Train Set\\ (distractors)\end{tabular} & \begin{tabular}[c]{@{}l@{}}Test Set\\ (distractors)\end{tabular} & F1 & Wt.-F1 \\ \midrule
 ResNet-18          & 3, 4 &  4 & 0.934 & 0.951             & & & 0, 1 &  2     &   0.667  &  0.741   \\
 Resnet-34          & 3, 4 &  4 & 0.934 & 0.951             & & & 0, 1 &  2     &   0.701  &  0.741 \\
 VGG*               & 3, 4 &  4 & 0.911 & 0.948             & & &  0, 1 &  2    &  0.720  &  0.590 \\
 DiscreteRel        & 3, 4 &  4 & 0.96  & 0.96                 & & &  0, 1 &  2    &  0.38  &  0.456 \\
 Real2Sim           & 3, 4 &  4 & 0.869 & 0.871             & & &  -  & -       &      N/A   &  N/A     \\
 Our Model (RN)     & 3, 4 &  4 & 0.949 & \textbf{0.970}    & & &  0, 1 &  2    &  0.649  &  0.720    \\
 Our Model (GNN)    & 3, 4 &  4 & 0.921 & 0.944             & & &  0, 1 &  2    &  0.841  &  \textbf{0.804} \\ \midrule
 ResNet-18          & 3, 4 &  6 & 0.802 & 0.771             & & &  0,1,2 &   3  & 0.844 &  0.902  \\
 ResNet-34          & 3, 4 &  6 & 0.823 & 0.833             & & &  0,1,2 &   3  & 0.777 &  0.871  \\
 VGG*               & 3, 4 &  6 & 0.667 & 0.481             & & &  0,1,2 &   3  & 0.788 & 0.842   \\
 DiscreteRel        & 3, 4 &  6 & 0.98  & \textbf{0.98}     & & &  0,1,2 &   3  &  0.200    & 0.658  \\
 Real2Sim           & 3, 4 &  6 & 0.960 & 0.960             & & &  - & -        & N/A &   N/A     \\
 Our Model (RN)     & 3, 4 &  6 & 0.971 & \textbf{0.981}    & & &  0,1,2 &   3  &  0.880 & 0.935   \\
 Our Model (GNN)    & 3, 4 &  6 & 0.948 & 0.969             & & &  0,1,2 &   3  & 0.921 &  \textbf{0.940}  \\ \midrule
 ResNet-18          & 3, 4 &  6 (diff. size) & 0.695 & 0.662 & & & 0,1,2  & 4   &  0.827 &  0.866   \\
 ResNet-34          & 3, 4 &  6 & 0.692 & 0.640             &  & & 0,1,2   & 4  &  0.833 & 0.868  \\
 VGG*               & 3, 4 &  6 & 0.640 & 0.601             & & & 0,1,2   & 4   & 0.720  & 0.770 \\
 DiscreteRel        & 3, 4 &  6 & 0.782 & 0.801             & & & 0,1,2   & 4   & 0.400  & 0.590 \\
 Real2Sim           & 3, 4 &  6 & 0.904 & 0.912             & & & 0,1,2   & 4   & N/A &  N/A \\
 Our Model (RN)     & 3, 4 &  6 & 0.952 & \textbf{0.952}    & & & 0,1,2   & 4   & 0.929 &  0.944  \\
 Our Model (GNN)    & 3, 4 &  6 & 0.952 & \textbf{0.952}    & & & 0,1,2   & 4   & 0.960 &  \textbf{0.960}  \\ \bottomrule 
\end{tabular}%
}
\caption{\footnotesize{Precondition learning results for sweeping objects in a line and food food skill.}}
\label{tab:results_objects_in_line_food_cut}
\vspace{-4mm}
\end{table}


\begin{figure}
    \begin{subfigure}{0.4\linewidth}
        \centering
        \includegraphics[width=0.8\linewidth]{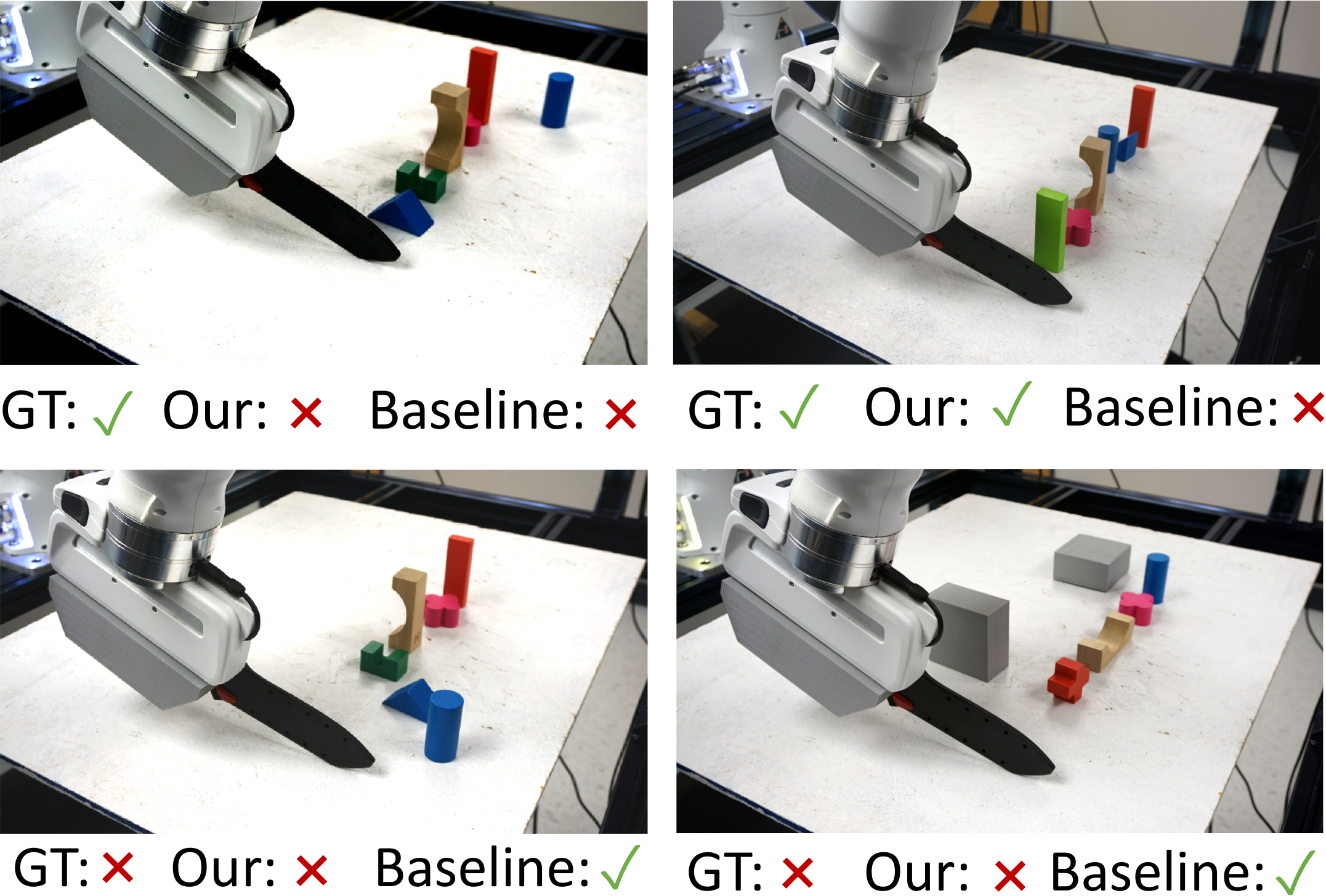}
    \end{subfigure}
    \begin{subfigure}{0.56\linewidth}
        \centering
        \includegraphics[width=1.0\linewidth]{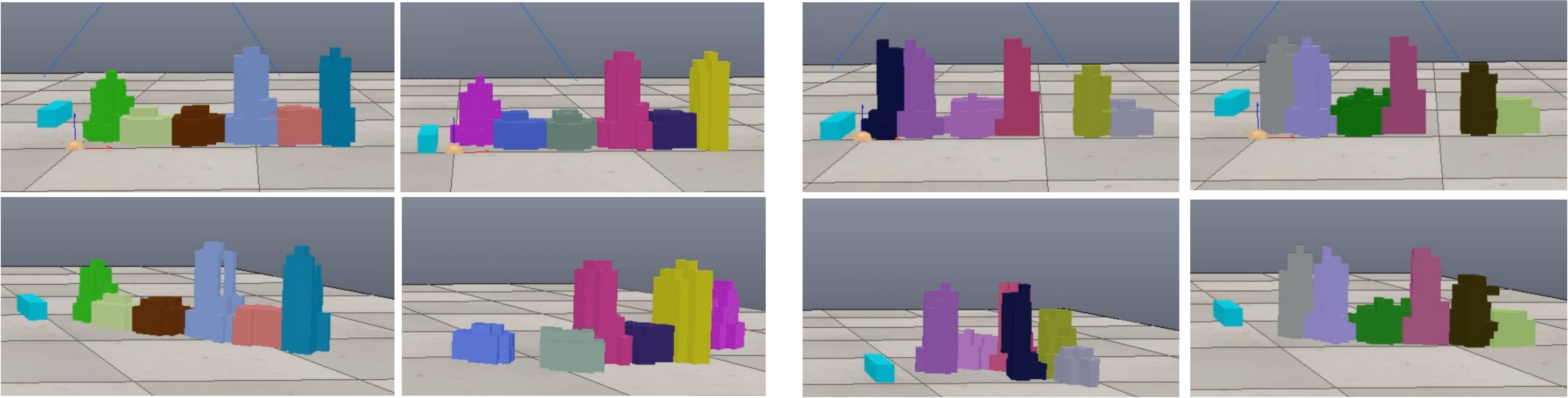}
    \end{subfigure}
    \caption{\footnotesize{\emph{Left:} Test set examples with \emph{different} shapes and sizes for sweeping task. Baseline here referrs to visual baseline. \emph{Right:} Example scenes for sweeping task when imported using the real2sim baseline. The top row shows the initial scene before the sweep while the bottom row shows the after scene.
    }}
    \label{fig:data_in_line_incorrect_6_diff_size_real2sim}
    \vspace{-4mm}
\end{figure}




\subsection{Sweeping Objects in Line}

Humans are adept at inferring if objects lie in an approximately straight line. 
However for robots, the presence of sensor noise, variances in object sizes and small amount of input data make this a challenging task.
Specifically, we look at task preconditions when a robot arm sweeps along the X-axis.
Figure~\ref{fig:data_in_line_cut_food_examples} (left) shows some examples from the train dataset. 
We collect scenes with variable number of objects (3, 4 and 6) and train on 25 different scenes.
Table~\ref{tab:results_objects_in_line_food_cut} (left) shows results for multiple train-test sets.
For the first train-test split all methods perform well. 
However, for the other splits \emph{i.e.}, when we test on 6 objects and objects with different shapes and sizes, the learning baseline methods perform poorly (wt. F1-score of $0.83$ and $0.66$).
This is not unexpected given that we have limited training data and a large mismatch between train and test distributions.  
While, discrete relations baseline performs quite well on the first two splits but poorly on the last split. 
Also, real2sim is unaffected by the distribution mismatch and performs quite well with a small decrease in its performance on the $3$rd split. 
Both of the latter baselines perform poorly due to more complex geometries of objects in the test set.
This leads to imperfect voxel representations,
which results in incorrect discrete relations and noisy simulation results when exported into the simulator respectively.
We visualise this in Figure~\ref{fig:data_in_line_incorrect_6_diff_size_real2sim} (Right) which shows two different sets of scenes,
each containing two scenes with similar initial configurations 
but very different final positions. 
Alternately, both of our models are able to outperform the baseline methods with a wt. F1-score of $0.96$ and $0.95$ respectively. 
Note that we only input object relation embeddings to our model and no other object specific information. 


\subsection{Food Cutting}

We use cutting food as our next task since it requires reasoning about multiple types of interactions.
For instance, to cut a food item the robot needs to hold the knife with its sharp edge in close contact with the food item, the knife should be oriented correctly to pass through the food item, and there should be no obstacle that can potentially hamper the back-and-forth cutting motion.
Since the simulator does not implement cutting we \emph{cannot} use the real2sim baseline.
Figures~\ref{fig:data_in_line_cut_food_examples} (right) shows some samples of the collected data.
Table~\ref{tab:results_objects_in_line_food_cut} (right) shows results for this task.
For the initial case, wherein we train with only one distractor object and test on more than one distractor objects, all methods perform poorly with a maximum wt. F1-score of our GNN based model $0.804$.
We believe this happens because with just one distractor in the train set many different object permutations were never observed and hence the model performs poorly on the test set.
However, when we train models with upto 2 distractors the performance on test sets with 3 or 4 distractors is much better. 
Also, for all of the train-test splits our relational precondition models are able to outperform all other baseline methods (wt. F1: $0.960$ and $0.940$).
Among our object relation based precondition models, the GNN based models perform slightly better. This can be attributed to their larger representation capacity, since we stack two GNN layers while the relation network model only contains one relational layer.


\begin{table}[]
\centering
\resizebox{0.9\textwidth}{!}{%
\begin{tabular}{@{}lllllllllll@{}}
\toprule
Model & Train Set & Test Set & F1 & Wt-F1 & \multicolumn{1}{l|}{} &  & Train Set & Test Set & F1 & Wt.-F1 \\ \midrule
ResNet-18                       &  3,4,5 & 7 & 0.741 & 0.836            &  &  & 3,4,5,7  & 6 & 0.770 & 0.820 \\
ResNet-34                       &  3,4,5 & 7 & 0.712 & 0.809            &  &  & 3,4,5,7  & 6 & 0.769 & 0.820 \\
VGG*                            &  3,4,5 & 7 & 0.711 & 0.804            &  &  & 3,4,5,7  & 6 & 0.744 & 0.818 \\
DiscreteRel                     &  3,4,5 & 7 & 0.561 & 0.655            &  &  & 3,4,5,7  & 6 & 0.594 &  0.617   \\
Real2Sim                        &  3,4,5 & 7 & 0.671 & 0.732            &  &  & 3,4,5,7  & 6 & 0.651 & 0.717 \\
Our Model (RN)                  &  3,4,5 & 7 & 0.685 & 0.775            &  &  & 3,4,5,7  & 6 & 0.679 & 0.771 \\
Our Model (GNN) all edges       &  3,4,5 & 7 & 0.825 & 0.869            &  &  & 3,4,5,7  & 6 & 0.819 & 0.857  \\
Our Model (GNN) sparse edges    &  3,4,5 & 7 & 0.864 & \textbf{0.898}   &  &  & 3,4,5,7  & 6 & 0.866 & \textbf{0.894} \\ \bottomrule
\end{tabular}%
}
\caption{\footnotesize{Results for block unstacking task with two different train-test splits.}}
\label{tab:results_block_stacking}
\vspace{-6mm}
\end{table}



\begin{table}[!htb]
    \begin{minipage}{.36\linewidth}
      \centering
        \resizebox{\textwidth}{!}{%
        \begin{tabular}{@{}lll@{}}
        \toprule
        Model     &  F1 & Wt-F1 \\ \midrule
        ResNet-18 &  0.695 &  0.719 \\
        ResNet-34 &  0.714 &  0.673 \\
        VGG*      &  0.545 & 0.665        \\
        DiscreteRel & 0.200 & 0.355        \\
        Real2Sim &  0.827 & 0.817 \\
        Our Model (RN) &  0.697 & 0.736    \\
        Our Model (GNN) all &  0.923 & \textbf{0.919}       \\
        Our Model (GNN) sparse &  0.923 & \textbf{0.919}       \\ \bottomrule \\
        \end{tabular}}
        \caption{\footnotesize{Results for precondition learning of box stacking task with completely different blocks (objects) in the test set.}}
        \label{tab:results_block_stacking_2}
    \end{minipage}%
    \quad
    \begin{minipage}{.6\linewidth}
      \centering
        \resizebox{\textwidth}{!}{%
        \begin{tabular}{@{}llll@{}}
        \toprule
        Model     & Task &  F1 & Wt-F1 \\ \midrule
        Only predictive loss & Cutting Food & 0.72 & 0.78    \\
        Only $\Delta p$ and $\Delta \theta$ predictive loss &  & 0.72 & 0.78    \\
        Only triplet loss  &  &0.828 & \textbf{0.866}       \\ \midrule
        Only predictive loss & Block Unstacking & 0.802 & 0.824    \\
        Only $\Delta p$ and $\Delta \theta$ predictive loss & no contacts & 0.775 & 0.800  \\
        Only triplet loss  & & 0.835 & \textbf{0.849}       \\ 
        Using mean position & & 0.681 & 0.721 \\ 
        Using mean position + bounding box & & 0.776 & 0.816 \\
        Only position loss (pred. + cont.) & & 0.672 & 0.759  \\ \bottomrule \\
        \end{tabular}}
        \caption{\footnotesize{Ablation results for different losses. The first three rows show values for food cutting task with 2 distractor objects in training and 4 the test set. The next 3 rows show block unstacking results with 3 to 5 objects in train set and 7 in the test set.}}
        \label{tab:results_ablation_1}
    \end{minipage} 
    \vspace{-6mm}
\end{table}

\begin{figure}
    \centering
    \begin{subfigure}{0.44\linewidth}
        \centering
        \includegraphics[width=1.0\linewidth]{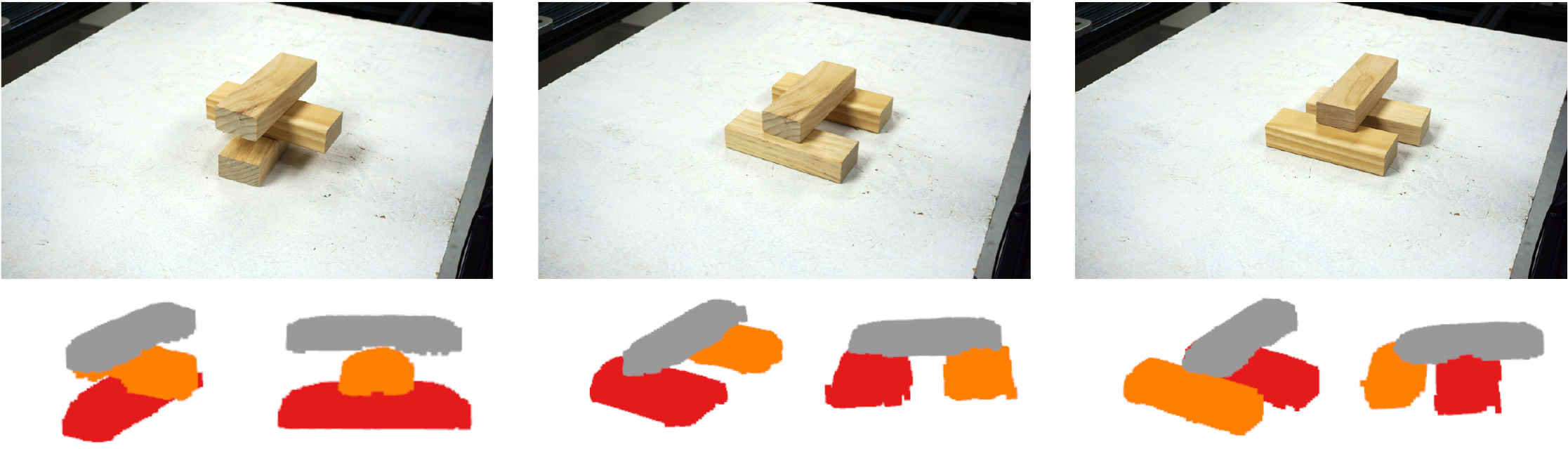}
    \end{subfigure}
    \quad
    \begin{subfigure}{0.50\linewidth}
        \centering
        \includegraphics[width=1.0\linewidth]{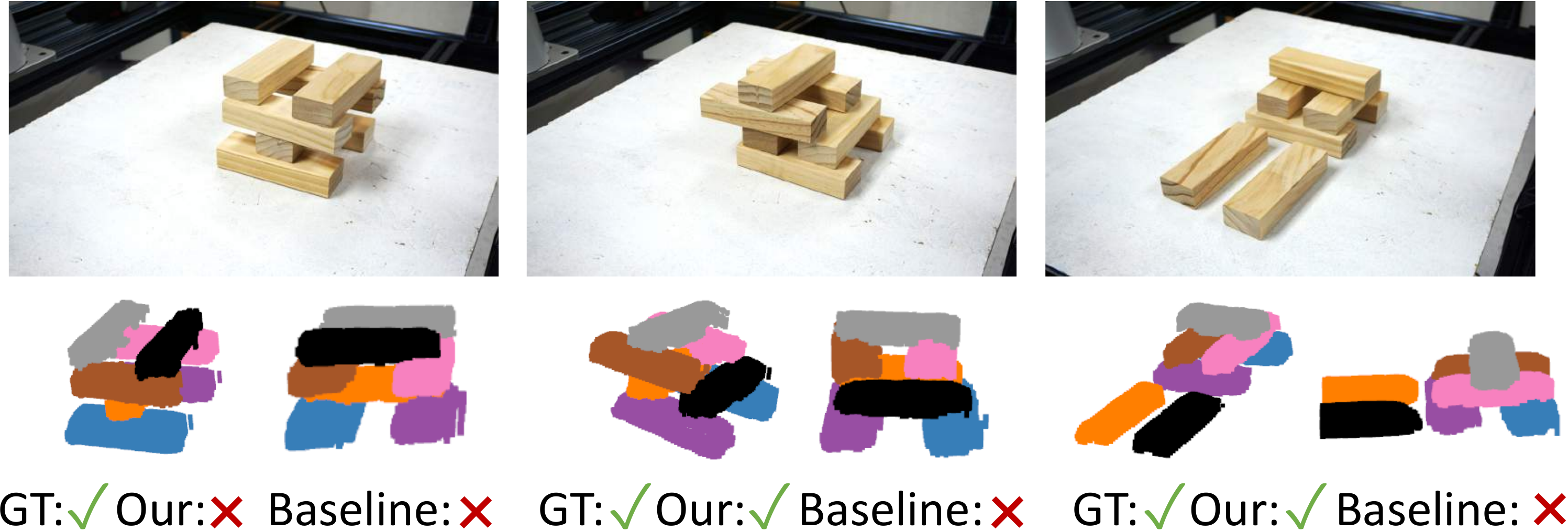}
    \end{subfigure}
    \caption{\footnotesize{\emph{Left:} Example scenes for block unstacking task. \emph{Right:} Test set examples for block unstacking task. The black block shows the block to be removed. Baseline here refers to visual baseline.
    }}
    \label{fig:data_block_stacking_3_blocks_test_7_blocks}
    \vspace{-4mm}
\end{figure}

\subsection{Block Unstacking}
\begin{figure}[t]
    \centering
    \begin{subfigure}{0.46\linewidth}
        \centering
        \includegraphics[width=\linewidth]{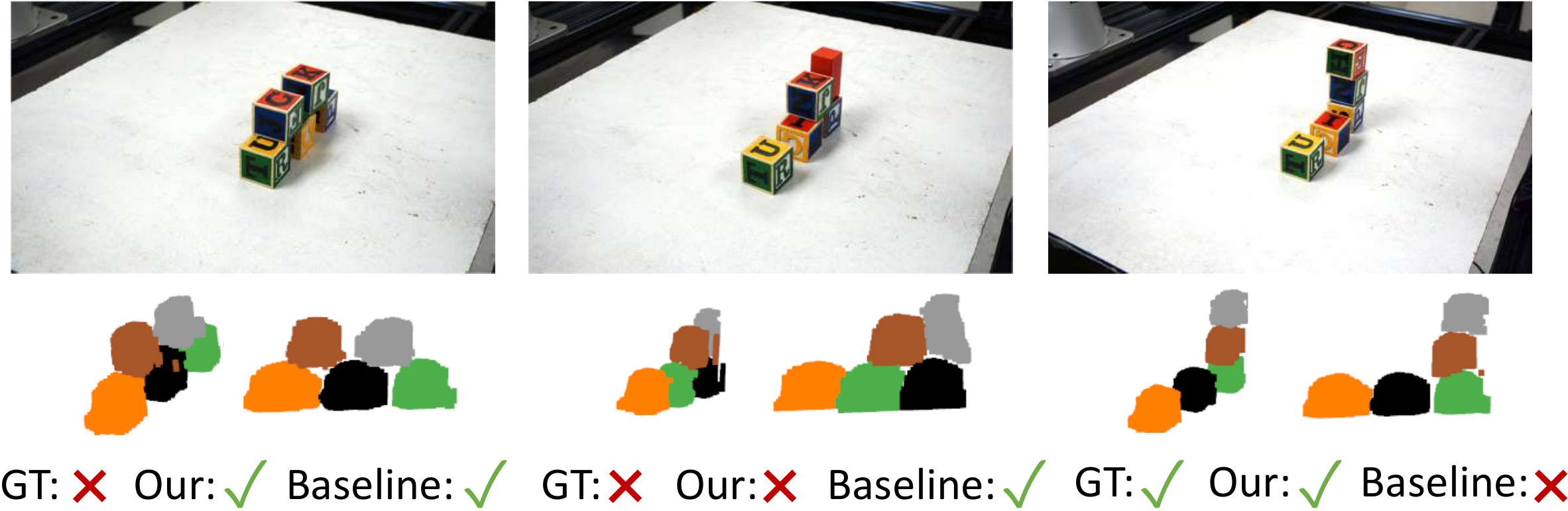}
    \end{subfigure}
    \quad
    \begin{subfigure}{0.50\linewidth}
        \centering
        \includegraphics[width=0.9\linewidth]{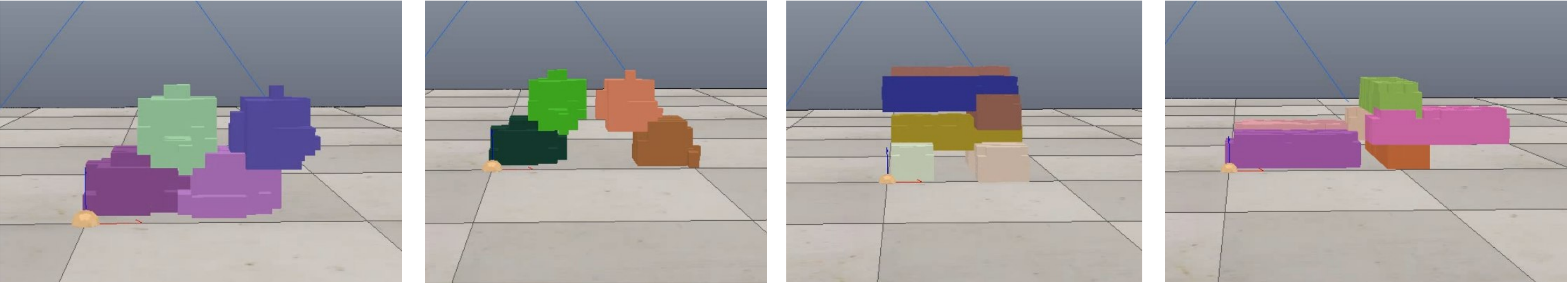}
    \end{subfigure}
    \caption{\footnotesize{\emph{Left:} Test set examples for block unstacking with \textbf{different shape and size} than train set. Baseline: visual baseline. \emph{Right:} Example scenes which are unstable but are predicted as stable by the Real2Sim baseline.}}
    \label{fig:data_block_stacking_test_5_square_real2sim_block_stack1}
    \vspace{-6mm}
\end{figure}

\vspace{-2mm}
Block unstacking involves a large amount of geometric and contact based reasoning since the blocks can be arranged in many complex configurations.
We formulate the precondition learning problem to predict the stability of a stack of blocks given a particular block to be removed. 
Figure~\ref{fig:data_block_stacking_3_blocks_test_7_blocks} (Left) shows examples with 3 blocks where the grey block can be safely removed from all scenes. 
In addition to the previous models, we add another GNN based model where graph edges only exist between blocks (vertices) that are closer than a certain threshold (0.1m), thus easing the learning problem.
Table~\ref{tab:results_block_stacking},~\ref{tab:results_block_stacking_2} shows results for multiple train-test splits. 
As seen above, our GNN based models outperform all other methods across all the different scenarios. 
The 3D CNN based baseline models have a higher wt. F1-score $(\sim 0.80)$ for the 1st-two splits but when tested on objects with different sizes (Table~\ref{tab:results_block_stacking_2}, Figure~\ref{fig:data_block_stacking_test_5_square_real2sim_block_stack1}(Left) ) they perform much worse. This shows that the 3D CNN based baseline models overfit to the train set objects and cannot transfer to objects with different shapes and sizes.
While the DiscreteRel baseline is clearly insufficient for this task.
For instance, given a 3-block configuration, with 1 block above and 2 supporting it from below. 
Although, this configuration has the same discrete representation, 
the overall stability of this 3-block configuration depends upon the location of the bottom blocks.
We discuss this in detail in 
Appendix~\ref{app:block_unstacking_detail}.

Also, the real2sim baseline performs poorly on the 1st-two test splits, wt. F1-score $(\sim 0.70)$. 
This is because blocks are in contact and quite close to each other, their voxel representations are quite noisy \emph{e.g.} with blocks often embedded into other blocks. 
These noisy voxel representations when imported into VREP lead to inaccurate predictions. Figure~\ref{fig:data_block_stacking_test_5_square_real2sim_block_stack1} (Right) visualizes scenes which are unstable but are predicted as stable in VREP. 
Interestingly, amongst our models, the relational network (RN) performs similar to the baseline models and much worse than the corresponding GNN models. 
We believe this is due to the complexity of the reasoning problem. 
Since we only use one layer of the relational network model, its representation capacity is much less than the corresponding GNN models with 2 layers. 
 
\section{Ablation Study}
\vspace{-2mm}


We perform an ablation study to understand the effects of different architecture and algorithmic choices in our proposed approach. First, we look at the effects of different loss functions used to train the object relation model. Table~\ref{tab:results_ablation_1} shows the effect of different loss functions on the precondition learning problem.
As seen above, using triplet loss performs better than predictive losses for both food cutting and block unstacking tasks. 
We believe this happens because the low dimensional embeddings learned using supervised losses alone might not be discriminatory enough for the downstream tasks. In comparison, since triplet loss explicitly forces embeddings to be further apart, it eases the learning problem for the precondition model.
Also, comparing Table~\ref{tab:results_ablation_1} with previous results we see that combining all the losses does outperform using any of the losses individually. 

To illustrate the utility of our learned embeddings, we evaluate only using the mean 3D position (similar to \cite{kroemer2016learning}) and bounding box of the blocks as input to our sparse GNN model.  
The bounding box is estimated min-max values for each axes from the voxel representation of the block. 
Table~\ref{tab:results_ablation_1} shows that using mean positions alone performs poorly (wt. F1-score: 0.72) while adding object bounds performs better (wt. F1-score:  0.816). 
However, our GNN model with learned object relations still outperforms them. 
One reason for this is the sensory noise in the input data, especially since object bounds are sensitive to outliers. 
More importantly, these results indicate the utility of our learned embeddings which perform well despite sensory noise and limited data.
Table~\ref{tab:results_ablation_1} also shows the utility of the contact based losses for the block unstacking task. Using contacts based losses increase the wt. F1-score on the block unstacking task from $0.80$ to $0.824$. This is not surprising since the block unstacking task requires inferring contact locations for stability prediction.


\section{Conclusion}
\vspace{-2mm}
We learn preconditions for manipulation skills by using structured scene representations that decompose scene into its objects and their relations. 
We propose a novel approach to learn generalizable continuous representations for these object relations.
Our approach has several advantages.
First, it is grounded in a robot's interaction with objects,
and hence we do not assume a fixed set of discrete object relations.
Second, simulations provides us access to large set of ground truth data such as contacts distribution which allow us to learn rich representations.
Finally, our approach can be directly used for 
precondition learning in a sample efficient manner.






\bibliographystyle{plainnat}
\acknowledgments{
This research was in part supported by NSF Award No. CMMI-1925130, the Office of Naval Research Grant No. N00014-18-1-2775, ARL grant W911NF-18-2-0218 as part of the A2I2 program, and by Sony AI.
}


\clearpage

\begin{appendices}


\section{Generating Pairwise Interactions in Simulation}

To position the referrant object in the scene we sample a location $(x, y, z)$ around the anchor object such that the distance between the anchor and referrant objects is less than 0.5 meter. 
We assume that the referrant object will be upright and hence to set its orientation we rotate it only around the Z-axis between $[-pi/6, pi/6]$ radians.  
To set the size of both the anchor and referrant objects we randomly sample each object dimension from $[0.04m, 0.20m]$. To save the voxel representation for each pairwise scene we set the voxel size to $0.01m$.
To perform the local perturbation actions we use a virtual robot which can move an object around. Our virtual robot consists of prismatic joints which allow it to move along the XYZ axis and revolute joints to rotate around them. 
We use a spring-damper system to control these joints. 
For the prismatic joints we set $K = 100 N/m$ and $C = 10 Ns/m$, while for revolute joints we use $K = 10 N/m$ and $C = 3 Ns/m$.

\section{Experimental Setup}

\begin{figure}[t]
    \centering
    \includegraphics[width=0.4\textwidth]{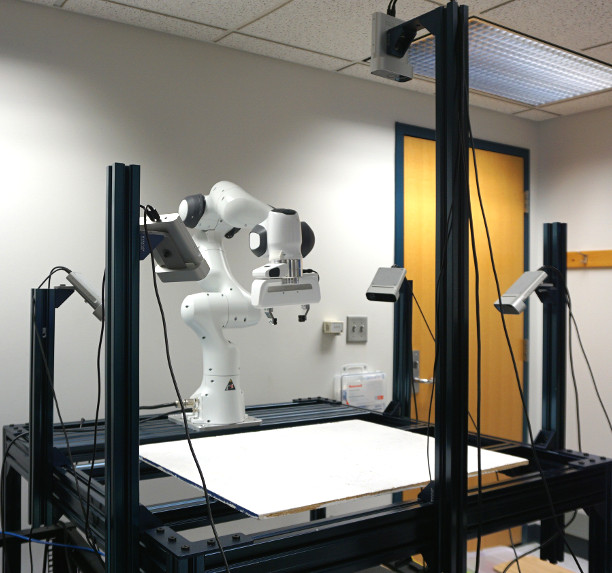}
    \caption{Experimental Setup with 5 cameras mounted around the robot arm. We combine the 3D point cloud output from each camera to create a full 3D scene.}
    \label{fig:exp_setup}
\end{figure}

Figure~\ref{fig:exp_setup} shows our setup to collect real world data and perform skills based on the learned precondition models. We have 5 cameras mounted near the robot's workspace that allow us to create a full 3D scene. To create the 3D scene we collect point cloud data from all the cameras and project them onto a common frame. We finally use volumetric tsdf integration \cite{curless1996volumetric} to fuse all of the point clouds together.

\section{Input Format}
\label{sec:input_format}
The input to our object relation mdoel and the baselines is of size $(C, 100, 100, 100)$, where $C$ is the number of channels. We next detail the exact input format for both the baseline and object relation model.

\subsection{Visual Baseline}
Since the input scene contains variable number of objects we use two different input formats for our baseline methods.
First, we only add masks for each object, i.e., each voxel is labeled with a value of 1 if it is occupied by any object, while unoccupied voxels are labeled with 0. Thus, this input format only requires one channel in its formulation.
For our second input format we add another channel to the input. This additional channel contains labels for each object in the scene. Thus, if a voxel is occupied by $i'th$ object we label it with integer $i$. This input format consists of 2 channels.
Additionally, we found using colored voxel representations to be extremely noisy since many objects in our experiments were quite close to each other.

We next discuss some changes to the input format for each specific task.
For the food cutting task we add labels that represent the knife object and the food object if they are present in the scene. For all other objects in the scene we add a separate distractor label.
For the block unstacking task we use two different labels. The block to be removed is marked by one label, while all the other objects in the scene use the other label. As before, we experiment with object ids for each block. But for the block unstacking task this method performed much worse and did not generalize at all when tested on a larger set of blocks.

\subsection{Object Relation Model}
The input to our object relational model $f_\text{rel}$ consists of 3 channels, where the first channel contains binary masks for both anchor and referrant objects. The second and third channel contains the binary masks for the anchor and referrant object separately.
When using the object relation model on real world manipulation data we take the object masks of two objects $i$ and $j$. We set $i$ as the anchor object and $j$ as the referrant object. 
We transform the view such that the anchor object $i$ is in the middle of the scene and centered at $(0, 0, 0)$. This is similar to the formulation used to train $f_\text{rel}$ in simulation.

\section{Baselines}
\subsection{DiscreteRel}
To get discrete relations directly using the input scenes with voxels we use an approach similar to \cite{zampogiannis2015learning}. 
We divide the space around each object (anchor) into multiple regions. 
These regions estimated using the bounds of the voxel objects.
Specifically, we use $26$ regions since a 3D object can be divided into $3^3$ regions and we do not include the middle region which contains the object.  
For every other object (referrant object) in the scene, we find the number of voxels of this referrant object which intersect with each of the above regions. 
The region with most voxels is used to represent the discrete relation between the anchor and referrant object.
Once we have the discrete relation for the objects, we use it in exactly the similar way as a continuous 
relation, \emph{i.e.}, we use the same network architectures for the precondition classification model.

\subsection{Real2Sim}
As shown in the main paper we evaluate our approach against the real2sim baseline. The real2sim baseline we use the same voxel representations as used by our approach and other baseline methods. 
As the first step to implement the real2sim baseline we need to import the voxel representations from the real world to VREP \cite{rohmer2013v}. 
Since these voxel representations do not conform to any fixed shape we cannot really import these object representations directly. 
Hence we import each voxel for every object separately into VREP. Thus each voxel is initially added as cuboid object. Once all the voxels of a given object are imported we group them together to form one composite object. Once all the objects in a scene have been imported we use it to test our preconditions. In addition to this, we also tried convex decomposition of the non-convex object shapes, but we found this to be incredibly slow and inaccurate for our purpose. 

Since the VREP simulator does not support cutting we use the sim2real baseline only for the sweeping objects in a line and block unstacking task.
To verify the preconditions for the sweeping objects in a line task we additionally add a virtual robot with a long cuboid attached to it. This robot consists of prismatic joints which allow us to move in the $XYZ$ axes. As before we use a spring-damper system to move the robot. At the beginning of the task we move the robot to its start position. We set the initial $X$ coordinate as the minimum $X$-position of all object corners minus some threshold $(0.02m)$. The $Y$ and $Z$ coordinates are set as the median of all object centers. This allows us to sweep through the scene in a consistent manner. To verify if the objects are indeed along a line, we evaluate if all the objects lie on one side (along the $X$-axes) of the robot handle.

We verify the preconditions for the block unstacking task by verifying if the blocks in the scene minus the block to be removed are stable or not. To achieve this we import all the blocks except the one to be removed into the simulator. Initially, all blocks are kept static. Once all of them have been imported we convert them to dynamic and allow them to interact with each other. To verify if the blocks are indeed stable we note each block's position before and after allowing dynamic interactions. If any of the blocks move by more than a certain specified threshold we assume that the blocks are not stable. 
To find the best threshold value we do a grid search in the set $(0.0005, 0.001, 0.0015, 0.002, 0.003, 0.004)m$. We found $0.0015$ and $0.002$ to give the best and very similar results.

\section{Architecture Details}

\subsection{Baselines}

For our baseline architecture we use the resnet model which has been adapted to operate on 3D input. We follow the architecture proposed in \cite{hara3dcnns}. 
However, we slightly tweak the ResNet architectures in \cite{hara3dcnns} for our purposes. 
For instance, we do not use the BatchNorm layers in the ResNet architectures since we empirically observed that they showed much worse performance. Additionally, in contrast to \cite{hara3dcnns} for the first convolution layer we use the same stride for all three input channels. Besides these changes, we follow the same architecture as \cite{hara3dcnns}. For other specific architecture details please look at Table~1 in \cite{hara3dcnns}. The output of the ResNet model is forwarded through 2 fully-connected layers with outputs of size 64 and 1 respectively. 

We also use another baseline similar in style to VGG \cite{simonyan2014very}, \emph{i.e.}, our model consists of a series of convolutions and ReLU non-linearities. More-specifically our architecture consists of 6 convolution layers, each of which operates on 3D input. Table~\ref{tab:vgg_params} lists the different parameters for each of these convolution layers. The output of each convolution layer is passed through ReLU non-linearity. The final output of the convolution is flattened to a 512 dimensional vector. This vector is then passed through multiple fully-connected layers with output size $[256, 64, 1]$.

\subsection{Precondition Learning Model}

\textbf{Relational Network:} We use the below form of relation network function, 
\begin{align}\label{eq:rel_network}
RN(S) = g_{\phi}\left( \sum_{i, j} f_{\theta}\left(f_\text{rel}(o_i, o_j) \oplus f_\text{rel}(o_j, o_i)\right) \right),
\end{align}
where $\oplus$ represents the concat operation, the function $f_\theta$ consists of two fully-connected (FC) layers with ReLU non-linearity in-between, and $g_\phi$ also consists of 2 FC layers with ReLU non-linearity. 

For our relation network model we implement the function $f_\theta$ using a two layer neural network with fully connected layers with outputs of size 128 and 32 respectively. 
We implement $g_\phi$, which acts as an accumulator, again by using two fully connected layers with outputs of size 16 and 1 respectively. 
We use the ReLU non-linearity for both $f_\theta$ and $g_\phi$.

\textbf{Graph Neural Network:} For our GNN model we use the following update model to process the output at each node,
\begin{align}\label{eq:gnn_1}
v^{(1)}_i = f_\psi\left(v_i \oplus \sum_{j}\left(f_\text{edge}(v^{(0)}_i, v^{(0)}_j, e_{ij})\right); \psi \right)
\end{align}
We add separate node and edge models for our GNN model. Our edge model takes the edge input and passes it through a fully connected layer with output size of 128, followed by ReLU, and finally with another fully connected layer which outputs a continuous embedding for this edge. These edge embeddings are used for the next graph layer as well as for each node being processed.
Our node model initially sums the edge embeddings for all the edges connected to this node. This sum of edge embeddings is then concatenated with the original node input and passed through a fully-connected layer with output size 128, followed by ReLU and then another fully conected layer with output size 128. This is similar in design to the edge update model.
We stack two layers of the above GNN model for precondition learning. The first layer uses an input of size 256 and outputs representations of size 128. The next layer uses an input of size 128 and outputs representations of size 128.
We sum the node and edge embeddings of the last GNN model and concatenate them together to form an input of size 256. This input is then passed through 2 fully conected layers with output size of 64 and 1 respectively.

\subsection{Object Relation Model}

Our object relation model initially uses the ResNet-18 model of \cite{hara3dcnns}. As before, we make two changes to this model, we use a stride of $(2, 2, 2)$ for the first convolutional layer, and we do not use the BatchNorm layers. The convolutional layer output is flattened into a vector of size $1536$ which is then projected through 2 fully-connected layers with output size $512$ and $256$. Thus, we get the final relational embedding of size $256$. We concatenate this with the action vector and pass them through three fully connected layers of size 128, 64 and 9 respectively. The final output of size 9 consists of the predicted position changes of size 3, predicted orientation changes of size 3 and predicted mean contact position of size 3.

\section{Training Details}

\begin{table}[t]
    \begin{minipage}{0.48\linewidth}
    \centering
    \resizebox{\textwidth}{!}{%
    \begin{tabular}{@{}lll@{}}
    \toprule
    Model                   & Batch Size & Learning Rates   \\ \midrule
    Object Relation         & 256        & 3e-4             \\
    RN based Precondition   & 8          & \{1e-3, 3e-4\}   \\
    GNN based Precondition  & 8          & \{1e-3, 3e-4\}   \\
    Baseline (Resnet-18/34) & 4, 8, 16   & {[}1e-3, 1e-4{]} \\
    VGG                     & 4, 8, 16   & {[}1e-3, 1e-4{]} \\ \bottomrule \\
    \end{tabular}}
    \caption{Batch Size and Learning rates for different models used}
    \label{tab:train_details}
    \end{minipage}
    \quad   
    \begin{minipage}{0.44\linewidth}
    \resizebox{\textwidth}{!}{%
    \begin{tabular}{@{}lll@{}}
    \toprule
            & conv-\{1,2\} & conv-\{3, 4, 5, 6\} \\ \midrule
    kernel  & 5            & 3            \\
    padding & (2, 2, 2)    & (2, 2, 2)    \\
    stride  & (1, 1, 1)    & (1, 1, 1)    \\ \bottomrule \\
    \end{tabular}}
    \caption{Parameters for the convolution layers of our VGG* network.}
    \label{tab:vgg_params}
    \end{minipage}
\end{table}

Table~\ref{tab:train_details} lists some of the training details for the different models used in our approach. Below we describe training details for each model separately.

\subsection{Object Relation Model}
As noted previously, for our embedding network we use a 3D CNN with a ResNet \cite{he2016deep} based architecture \cite{hara3dcnns}, specifically we use the ResNet-18 architecture. The output of the ResNet18 model is then downsampled using a linear layer to a size of 256. 
To train the above embedding model, we use the Adam optimizer \cite{kingma2014adam} and set an initial learning rate of $3e-4$ and a batch size of 256. 
Our overall loss function for the

The total loss for the object relation model can be written as,
\begin{align*}
    \mathcal{L} &= \lambda_\text{pos}^\text{cont}\times \mathcal{L}_\text{pos}^\text{cont} + \lambda_\text{orient}^\text{cont}\times \mathcal{L}_\text{orient}^\text{cont} \\
    &+ \lambda_\text{pred} \times \mathcal{L}_\text{pred} + \lambda_\text{orient} \times \mathcal{L}_\text{orient} + \lambda_\text{contact} \times \mathcal{L}_\text{contact},
\end{align*}
where the first set of losses $\mathcal{L}_\text{pos}^\text{cont}$, $\mathcal{L}_\text{orient}^\text{cont}$ are the position and orientation based contrastive losses. 
The next set of losses $\mathcal{L}_\text{pred}$, $\mathcal{L}_\text{orient}$, $\mathcal{L}_\text{contact}$ are the direct supervised losses. 
When training the object relation model with all the above losses we set, $\lambda_\text{pos}^\text{cont} = 2$, $\lambda_\text{orient}^\text{cont} = 2$. For the supervised losses we set $\lambda_\text{pos} = 1$, since orientation values change less we set $\lambda_\text{orient} = 10$, and finally $\lambda_\text{contact} = 1$.
Alternatively, when training the model with just contrastive losses we set $\lambda_\text{pos}^\text{cont} = \lambda_\text{orient}^\text{cont} = 1$.

\begin{table}[]
\centering
\begin{tabular}{@{}ll@{}}
\toprule
Parameter              & Value \\ \midrule
$\Delta p^r_\text{sim}$, $\Delta p^r_\text{diff}$ & 0.2, 0.21   \\
$\Delta \theta_\text{sim}$, $\Delta \theta_\text{diff}$ & 0.004, 0.008   \\
$\gamma$ for $\Delta p$ & 2.0   \\
$\gamma$ for $\Delta \theta$         & 2.0   \\ \bottomrule \\
\end{tabular}
\caption{Different hyper-parameters for contrastive loss formulation.}
\label{tab:contrastive_loss_params}
\end{table}

\textbf{Contrastive Loss:}
To implement the contrastive loss we use a batch all strategy, \emph{i.e.}, instead of explicitly sampling anchor, positive and negative pairs separately, we sample a batch of data and compare all scene triplets. 
We compare normal and adaptive actions for each triplet pair separately, \emph{i.e.}, for each scene in the triplet we compare their adaptive action effects with the other scene's adaptive action effects only. 
To compare normal actions with different action magnitudes we use a threshold of $0.04m$ to classify similar actions. Thus, scenes with difference in action magnitude greater than the above threshold are not compared directly in contrastive losses.

As noted in the paper, we use the following to compare the action effects on position changes,
\begin{align}
    \Delta p^r = \frac{\Delta p^{\text{observed}}}{\Delta p^{\text{desired}}},
\end{align}
where $\Delta p^{\text{observed}}$ is the observed change in referrant object center position, and $\Delta p^{\text{desired}}$ is the desired change i.e. action vector. 
We set $\Delta p^{\text{desired}}$ to 1 for normal actions, while for adaptive actions it is set to the voxel distance between objects centers. 
To find scenes with similar action effects we use a threshold of $0.2$. More precisely, given two scenes ($m$ and $n$) and their action effects in terms of $\Delta p^r$, we say these two scenes are similar if their action effects are  below the threshold for all actions, $|\Delta p^r_m - \Delta p^r_n| < 0.2$. Analogously, if $|\Delta p^r_m - \Delta p^r_n| > 0.21$ we set these two scenes as dissimilar.
For orientation changes we set $\Delta \theta_\text{sim} = 0.004$ and $\Delta \theta_\text{diff} = 0.008$. 
Finally, we set the contrastive loss margin $\gamma = 2.0$ for both position and orientation changes.
Table~\ref{tab:contrastive_loss_params} lists the above parameters for contrastive losses.


\begin{figure}
    \centering
    \begin{subfigure}{0.4\linewidth}
        \centering
        \includegraphics[width=0.9\linewidth]{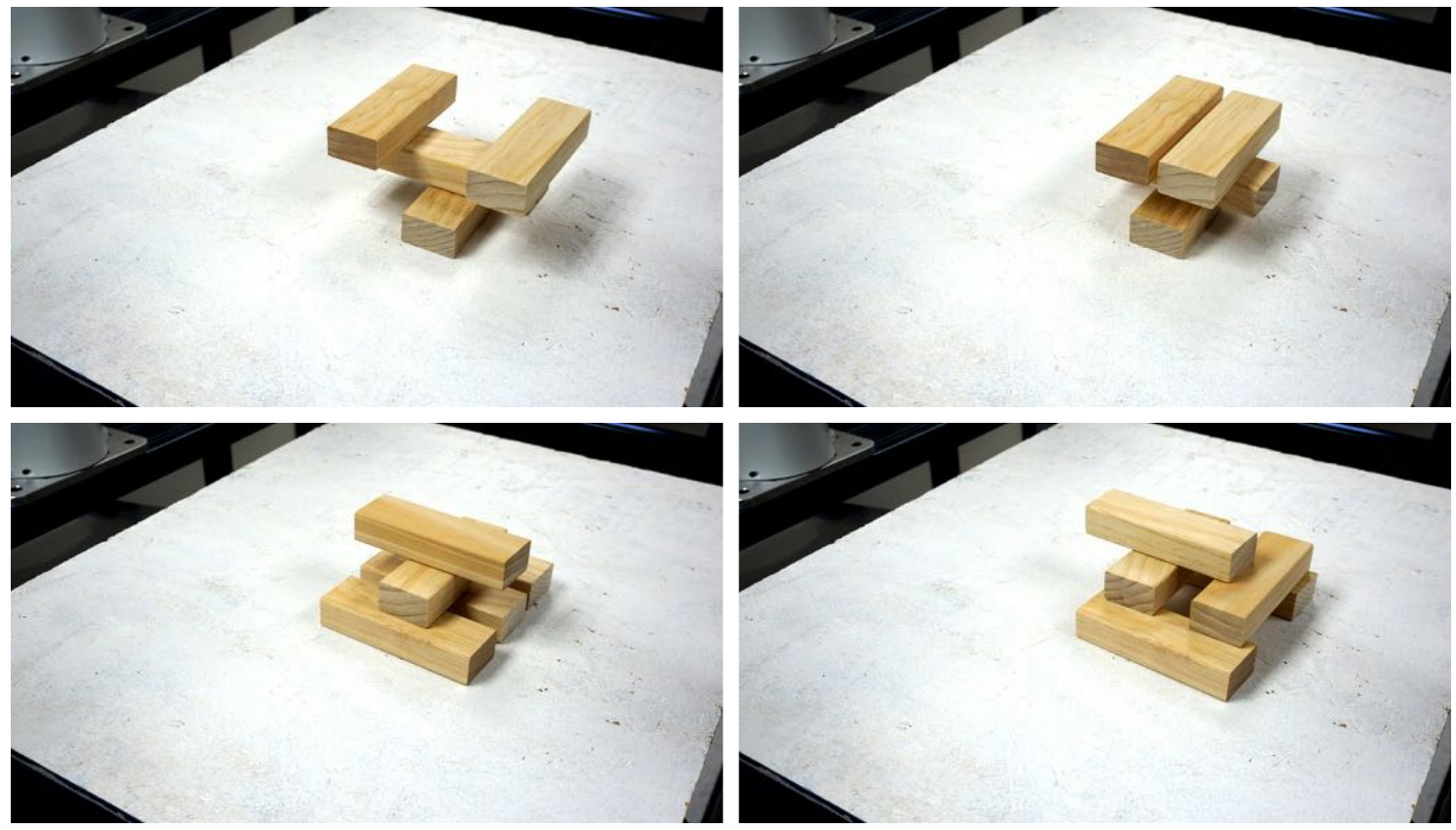}
    \end{subfigure}
    \quad
    \begin{subfigure}{0.46\linewidth}
        \centering
        \includegraphics[width=\linewidth]{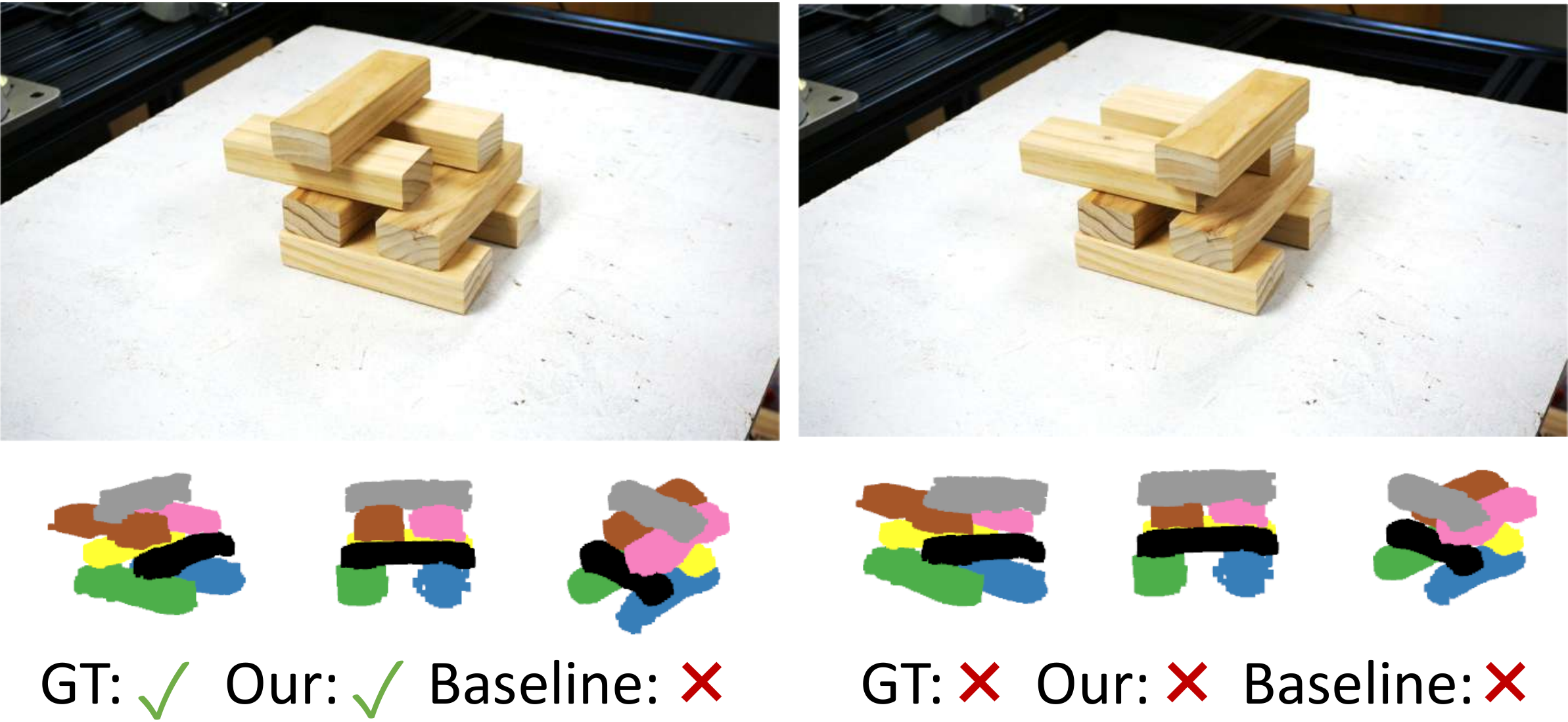}
    \end{subfigure}
    \caption{\emph{Left:}: Example train scenes with 4 and 5 blocks used in the block unstacking task. \emph{Right:} Sample test scenes which are correctly classified by our model but incorrectly by the learning based baselines. In both images the top block is at different locations.}
    \label{fig:data_block_stacking_train_4_5_test_7}
\end{figure}

\begin{table}[]
\centering
\resizebox{0.7\textwidth}{!}{%
\begin{tabular}{@{}lllll@{}}
\toprule
Model     & Train Set    & Test Set  & F1 & Wt-F1 \\ \midrule
ResNet-18 & 0, 1, 2 distractors & 3 distractors & 0.844     & 0.902 \\
ResNet-34 & 0, 1, 2 & 3 & 0.877     & 0.871 \\
VGG*      & 0, 1, 2 & 3 & 0.788     & 0.842 \\
DiscreteRel      & 0, 1, 2 & 3 & 0.601     & 0.658 \\
Real2Sim & - & - & N/A & N/A \\
Our Model (RN) & 0, 1, 2 & 3 & 0.880     & 0.935        \\
Our Model (GNN) & 0, 1, 2 & 3 & \textbf{0.921}     & \textbf{0.940}        \\ \bottomrule \\

\end{tabular}}
\caption{Precondition learning results for cutting food skill.}
\label{tab:results_food_cutting}
\end{table}

\begin{table}[t]
\centering
\resizebox{0.7\textwidth}{!}{%
\begin{tabular}{@{}lllll@{}}
\toprule
Model     & Train Set    & Test Set  & F1 & Wt-F1 \\ \midrule
ResNet-18 & 6, 7 & 4 & 0.735     & 0.765       \\
ResNet-34 & 6, 7 & 4 & 0.765     & 0.784       \\
VGG*      & 6, 7 & 4 & 0.688     & 0.712        \\
DiscreteRel & 6, 7 & 4 & 0.647 & 0.621 \\
Real2Sim  & -  & 4 & \textbf{0.89} & \textbf{0.86} \\
Our Model (RN) & 6, 7 & 4 & 0.74     & 0.764        \\
Our Model (GNN) all edges & 6, 7 & 4 & 0.772     & 0.798        \\ 
Our Model (GNN) sparse edges & 6, 7 & 4 & 0.824 & 0.825        \\ \bottomrule \\
\end{tabular}}
\caption{Results for precondition learning of block unstacking task with 6 and 7 objects in the train set and 4 objects in the test set.}
\label{tab:results_block_stacking}
\end{table}

\subsection{Precondition Learning Models}

\textbf{Our Model:} For our RN and GNN based models, we use the Adam optimizer \cite{kingma2014adam} and set an initial learning rate of $3e-4$. For accelerated learning, we also test with a larger learning rate of $1e-3$. 
Since the precondition model is trained on less data we use a smaller batch of size 8 only. We also decay the learning rate with 0.995 after every epoch.

\textbf{Baselines:}
For the baseline models we use the Adam optimizer as well. We experiment with both random initialization as well as kaiming initialization for the resnet based baseline methods \cite{he2016deep}. 
We do a grid search to find the best learning rate between $[1e-3, 1e-4]$ and report numbers with it. 
Also, given the small train data size we test with multiple batch sizes of 4, 8 and 16 and report numbers with the best results.

Table~\ref{tab:train_details} lists the different parameters used to train the precondition models.

\section{Task Setup}

\textbf{Cutting Food: } For the food cutting data we use a custom 3D printed tool holder with an embedded knife that the robot can grasp and use to cut food items. 
We use multiple different target food items and obstacles with different shapes and sizes. 
Unlike the sweeping objects in a line task we also add knife and food labels to their respective voxel representations, while all the other objects have no label. 
We create scenes with food, knife and multiple objects with a maximum of upto 6 objects in the scene.

\textbf{Block Unstacking:} To create the dataset we pre-program the robot to assemble a block of stacks in different configurations and then manually label which blocks are crucial for stability. 
We create scenes with variable number of blocks from $3$ to $7$. For each set of blocks we create between $10$ to $20$ scenes with different configurations. 
In addition to the previous models, we add another GNN based model where graph edges only exist between blocks (vertices) that are closer than a certain threshold (0.1m). This allows us to remove edges between blocks which are far apart and thus eases the learning problem.

\begin{figure}
    \centering
    \includegraphics[width=0.8\linewidth]{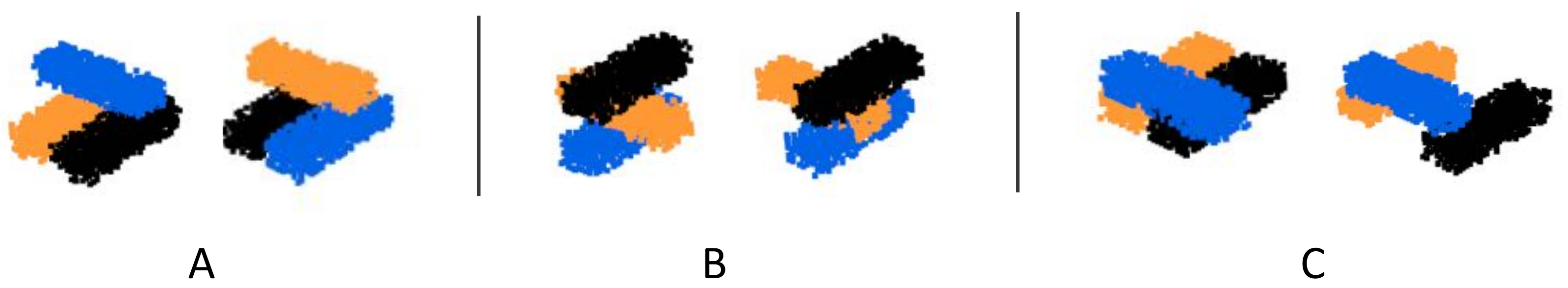}
    \caption{Multiple different pairs of scenes containing 3 objects, where discrete relations are insufficient o predict the stability of the scene. 
    For each scene the \emph{black} block is the block to be removed.}
\label{fig:results_discrete_relation_fail}
\end{figure}

\section{Additional Results}

In this section we discuss some additional results for our main experiments.

\subsection{Cutting Food}

Table~\ref{tab:results_food_cutting} shows the results for the food cutting experiment when we have 3 distractor objects in the test set, but only upto 2 distractor objects in the train set.
As observed above, the GNN based model performs the best with a wt. F1-score of $0.94$. Additionally, the baseline models also perform well with a maximum wt. F1-score of $0.902$. 
The good performance of the baseline models on food cutting task can be attributed to the fact that there exist only two main objects in the scene (food and the knife).
Since we provide separate labels to both of these objects, the baseline models only needs to focus on these objects and hence does not require complex compositional reasoning. 
Thus, even with increasing number of distractor objects, the baseline model still performs quite well.

\subsection{Block Unstacking}
\label{app:block_unstacking_detail}

For the block unstacking task we also experiment with the scenario where the train set contains larger number of objects as compared to the test set. Table~\ref{tab:results_block_stacking} shows results when the train set contains 6 or 7 objects while the test set contains only 4 objects. 
Interestingly, the performance of our GNN based models also reduces in this scenario. We get a max. wt. F1-score of $0.825$ only, which is worse as compared to the previous train-test splits. However, our GNN based models still perform better than the visual baseline models, whose maximum wt F1 score is $0.784$.
This result indicates that although structured representations are extremely useful, we cannot assume that training on a large set of objects will automatically generalize to fewer objects as well. The real2sim baseline performs the best (wt. F1 score: $0.889$) in this setting. This is because using 4 blocks, we can only create a limited number and types of block stacks. Hence, given only stacks with 3 blocks (since we remove 1 block before exporting into the simulator) the Real2Sim baseline is able to perform quite well.
Additionally, a small number of blocks also result in fewer perception errors, these errors when transferred to simulation do not affect the precondition output significantly. 
This is also similar to the real2sim results for sweeping task (main paper) wherein we observe that larger number of objects lead to worse performance.

Figure~\ref{fig:data_block_stacking_train_4_5_test_7} (Right) shows example scenes which show the complex reasoning required for solving the block unstacking task.
The only difference between the two scenes in Figure~\ref{fig:data_block_stacking_train_4_5_test_7} (Right) is in the position of the top block which results in different ground truth precondition labels.  While our precondition model is correctly able to predict the preconditions in both scenes, the visual baseline model always predicts false. The above example clearly shows that our model is able to perform complex reasoning using the learned object relational embeddings.

\textbf{Why DiscreteRel baseline performs poorly?}
We also look at why discrete relations are often insufficient to model relations for manipulation tasks. 
To show this we use the block unstacking task, in the simplest possible setting -- with only 3 objects in the scene.
Figure~\ref{fig:results_discrete_relation_fail} plots multiple different pairs (A, B, C) of such scenes.
Each of these scenes are part of the train set with 3 objects.
Also, for each scene pair the extracted discrete relations are similar (aside from image A, which shows the same scene).
This is because each referrant object occupies relatively similar amount of space around every anchor object.
However, despite similar discrete relations, 
both scenes in each scene pair (A, B, C) have different precondition outputs (when the black block is removed).
For instance, in Figure~\ref{fig:results_discrete_relation_fail} (B) 
removing the black box from left figure (scene) results in a stable block configuration.
While for the right scene in Figure~\ref{fig:results_discrete_relation_fail} (B),
this results in the orange block falling and hence not a stable configuration.
These results show the inability of discrete relations 
to differentiate between seemingly similar scenes but 
with very object arrangements.
Finally, each of the above scene only contains 3 objects. 
With increasing number of objects in the scenes, the ambiguity associated with discrete relations increases further,
which results in an overall poor performance across all different task configurations.

\end{appendices}

\end{document}